\documentclass[letterpaper, 10pt, conference]{ieeeconf}
\IEEEoverridecommandlockouts
\usepackage[english]{babel}

\usepackage{amsfonts, amssymb, amsmath, amsthm}
\usepackage{bm}
\usepackage{mathtools}
\usepackage{cite}
\usepackage{graphicx}
\usepackage{subcaption}
\usepackage{booktabs}
\usepackage{multirow}
\usepackage{algorithm}
\usepackage{algpseudocode}
\usepackage{cancel}
\usepackage{tikz}
\usepackage[bookmarks=true, breaklinks=true, hidelinks]{hyperref}

\newcommand{\calD}{{\cal D}}

\newcommand{\calL}{{\cal L}}

\newcommand{\calN}{{\cal N}}

\newcommand{\bfa}{\mathbf{a}}
\newcommand{\bfb}{\mathbf{b}}
\newcommand{\bfc}{\mathbf{c}}

\newcommand{\bfe}{\mathbf{e}}
\newcommand{\bff}{\mathbf{f}}
\newcommand{\bfg}{\mathbf{g}}
\newcommand{\bfh}{\mathbf{h}}

\newcommand{\bfk}{\mathbf{k}}

\newcommand{\bfp}{\mathbf{p}}
\newcommand{\bfq}{\mathbf{q}}
\newcommand{\bfr}{\mathbf{r}}

\newcommand{\bfu}{\mathbf{u}}
\newcommand{\bfv}{\mathbf{v}}

\newcommand{\bfx}{\mathbf{x}}

\newcommand{\bfz}{\mathbf{z}}

\newcommand{\bfepsilon}{\boldsymbol{\epsilon}}

\newcommand{\bfeta}{\boldsymbol{\eta}}
\newcommand{\bftheta}{\boldsymbol{\theta}}

\newcommand{\bfpi}{\boldsymbol{\pi}}

\newcommand{\bftau}{\boldsymbol{\tau}}

\newcommand{\bfphi}{\boldsymbol{\phi}}

\newcommand{\bfomega}{\boldsymbol{\omega}}

\newcommand{\bfell}{\boldsymbol{\ell}}

\newcommand{\bfA}{\mathbf{A}}
\newcommand{\bfB}{\mathbf{B}}

\newcommand{\bfD}{\mathbf{D}}
\newcommand{\bfE}{\mathbf{E}}
\newcommand{\bfF}{\mathbf{F}}

\newcommand{\bfH}{\mathbf{H}}
\newcommand{\bfI}{\mathbf{I}}
\newcommand{\bfJ}{\mathbf{J}}
\newcommand{\bfK}{\mathbf{K}}

\newcommand{\bfM}{\mathbf{M}}

\newcommand{\bfP}{\mathbf{P}}
\newcommand{\bfQ}{\mathbf{Q}}
\newcommand{\bfR}{\mathbf{R}}

\newcommand{\bfU}{\mathbf{U}}
\newcommand{\bfV}{\mathbf{V}}
\newcommand{\bfW}{\mathbf{W}}

\newcommand{\bfSigma}{\boldsymbol{\Sigma}}

\newcommand{\bbH}{\mathbb{H}}

\newcommand{\bbR}{\mathbb{R}}

\theoremstyle{definition}

\DeclareMathOperator*{\argmin}{arg\,min}

\newcommand{\NEW}[1]{{\color{black}#1}}

\begin{document}

\title{\textbf{Adapting Neural Robot Dynamics on the Fly for Predictive Control}}
\author{Abdullah Altawaitan and Nikolay Atanasov%
\thanks{We gratefully acknowledge support from NSF CCF-2112665 (TILOS).}%
\thanks{The authors are with the Department of Electrical and Computer Engineering, University of California San Diego, La Jolla, CA 92093, USA, e-mails: {\tt\small \{aaltawaitan,\allowbreak natanasov\}@ucsd.edu}. A. Altawaitan is also affiliated with Kuwait University as a holder of a scholarship.}
}
\maketitle
\thispagestyle{empty}
\pagestyle{empty}

\begin{abstract}
Accurate dynamics models are critical for the design of \NEW{predictive controllers} for autonomous mobile robots. Physics-based models are often too simple to capture relevant real-world effects, while data-driven models are data-intensive and slow to train. We introduce an approach for fast adaptation of neural robot dynamic models that combines offline training with efficient online updates. Our approach learns an incremental neural dynamics model offline and performs low-rank second-order parameter adaptation online, enabling rapid updates without full retraining. We demonstrate the approach on a real quadrotor robot, achieving robust predictive tracking control in novel operational conditions.
\end{abstract}

\section*{SUPPLEMENTARY MATERIAL}
Software and videos supplementing this paper: 

\centerline{\url{https://altwaitan.github.io/on_the_fly}}

\section{INTRODUCTION}

Autonomous mobile robots operating in dynamic environments benefit from accurate dynamics models to ensure high-performance and constraint satisfaction in low-level control. Nominal physics-based models degrade under changing conditions such as disturbances, payload variations, and motor wear, leading to poor control performance and potential constraint violation and instability. An alternative considering training dynamics models from data offline is impractical when conditions change during operation. This motivates the need for online model adaptation from onboard sensor measurements such that model accuracy is improved compared to white-box physics-based models and \NEW{model adaptation} is faster compared to black-box learned models.

Adaptive control \cite{krstic1995nonlinear, ioannou1996robust} offers well-established tools to estimate and compensate for disturbances and parameter variations online. Commonly, uncertainties are represented via predefined nonlinear basis functions with unknown weights, which are updated through adaptation laws grounded in Lyapunov stability theory \cite{krstic1995nonlinear}, sliding-mode methods \cite{slotine1991applied}, $\calL_{1}$ adaptation \cite{hovakimyan2010l1}, and model reference adaptive control \cite{ioannou1996robust}. These methods typically parameterize uncertainties as additive disturbances with a prescribed structure, which limits their ability to capture complex, state-dependent dynamics changes. 

Machine learning offers a data-driven alternative, as neural networks can approximate unmodeled dynamics without requiring a predefined parametric form \cite{richards2023control, shi2021meta, bisheban2020geometric}. In recent years, meta-learning methods \cite{finn2017model, rajeswaran2019meta, o2022neural} have emerged as a powerful tool to adapt prior learned models quickly to new tasks or changing conditions. Specific to aerial robot applications, meta-learning adaptation has been demonstrated to compensate for wind disturbances \cite{o2022neural, he2025self}, and capture the impact of suspended payloads \cite{belkhale2021model, sanghvi2025occam} or aerodynamic effects at high-speeds \cite{wei2025meta}. Rather than learning dynamics models, other works directly learn adaptive control policies \cite{zhang2025learning} or controller parameters \cite{sanghvi2025occam}. A common requirement of these approaches is a large, diverse dataset to encode shared structure, and training is computationally expensive due to second-order gradient computation.

\begin{figure}[t]
    \centering
    \includegraphics[width=\linewidth]{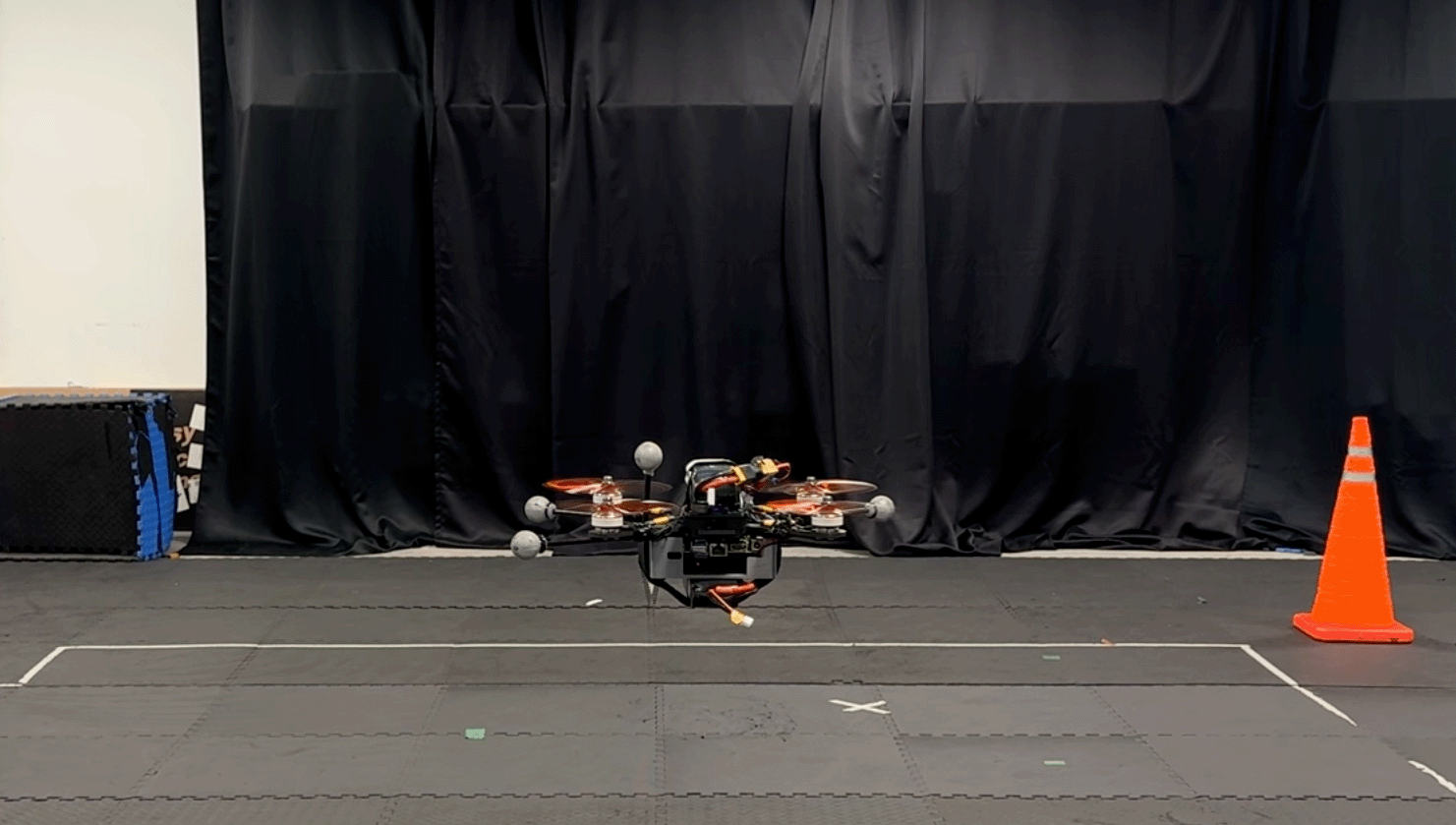}
    \caption{Quadrotor robot adapting to a payload equal to $35\%$ of its weight on the fly while tracking a reference trajectory.}
    \label{fig:thumbnail}
\end{figure}

While the above approaches leverage offline diversity for rapid online adaptation, the continuous and active updating of a learned dynamics model in real-time during deployment remains challenging. Saviolo et al.~\cite{saviolo2023active} take a step in this direction by learning the dynamics offline and adapting the last network layer online via stochastic gradient descent over one-step forward predictions for MPC. The method is limited to off-board execution due to its computational cost and relies on a low-level PID controller that abstracts away physical parameters such as inertia and motor characteristics.  

We develop an approach for on-the-fly neural dynamics learning and predictive control that adapts a pre-trained dynamics model in real-time on-board the system. While powerful neural network models have been developed for robot dynamics \cite{xu2025neural}, few have been deployed for online adaptation due to the cost of re-training. Low-rank adaptation offers a promising direction, having proven effective for fine-tuning large language \cite{hu2022lora} and vision-language-action models \cite{kim2024openvla}. We show that it also enables fast on-board dynamics model adaptation on an embedded, resource-constrained, CPU-only platform. Our contributions are: 
\begin{itemize}
    \item an incremental neural dynamics model trained offline via multi-step rollouts,
    \item a low-rank second-order parameter adaptation algorithm for real-time, on-board model updates, 
    \item demonstration of predictive trajectory tracking control with online model adaptation under changing conditions on a real quadrotor robot.
\end{itemize}

\section{PRELIMINARIES}
This section introduces background material that will be used throughout the paper.

\subsection{Notation}
We use bold lowercase symbols, $\bfa \in \mathbb{R}^{n}$, for vectors and bold uppercase symbols, $\bfA \in \mathbb{R}^{n \times m}$, for matrices. We use superscripts to denote derivatives. For example, $(\cdot)^{\bfx}$ and $(\cdot)^{\bfx \bfx}$ denote the gradient and Hessian with respect to $\bfx$, respectively. The weighted Euclidean norm is defined as $\| \bfx \|_{\bfW}^{2} := \bfx^{\top} \bfW \bfx$.

\subsection{Quaternions}
We represent orientations with unit quaternions, $\bfq \in \bbH_1$, which offer a compact representation with an $\mathbb{R}^{4}$ vector and one unit norm constraint. The group of unit quaternions is defined as  
\begin{align}
    \bbH_1 := \{ \bfq \in \mathbb{R}^{4} : \| \bfq \| = 1  \},
\end{align}
where $\bfq = \left( q_w , \bfq_{v} \right)$ consists of a scalar $q_w \in \mathbb{R}$ and a vector $\bfq_{v} \in \mathbb{R}^{3}$. Quaternion multiplication satisfies 
\begin{align}
    \bfq_{1} \otimes \bfq_{2} \triangleq \left[ \bfq_{1} \right]_{L} \bfq_{2} = \left[ \bfq_{2} \right]_{R} \bfq_{1}, 
\end{align}
with $\left[ \ \cdot \ \right]_{L}$ and $\left[ \ \cdot \ \right]_{R}$ denoting the left and right quaternion product matrices, respectively, defined as 
\begin{align}
    \left[ \bfq \right]_{L} = 
    \begin{bmatrix}
        q_w & -\bfq_{v}^{\top} \\ 
        \bfq_{v} & q_w \bfI_3 + \left[ \bfq_{v} \right]_{\times} 
    \end{bmatrix}, 
    \left[ \bfq \right]_{R} = 
    \begin{bmatrix}
        q_w & -\bfq_{v}^{\top} \\ 
        \bfq_{v} & q_w \bfI_3 - \left[ \bfq_{v} \right]_{\times}. 
    \end{bmatrix}. \notag
\end{align}
The operator $\left[ \ \cdot \ \right]_{\times}: \mathbb{R}^{3} \rightarrow \mathfrak{so}(3)$ maps a vector in $\mathbb{R}^{3}$ to a $3 \times 3$ skew-symmetric matrix. The inverse of a quaternion $\bfq^{-1}$ is $\left( q_w , -\bfq_{v} \right)$. A vector $\bfv \in \mathbb{R}^{3}$ is rotated by a quaternion $\bfq$ as 
\begin{align}
    \bfq \cdot \bfv \triangleq
    \bfq \otimes \bfH \bfv \otimes \bfq^{-1} = \left[ \bfq \right]_{L} \left[ \bfq \right]_{R}^{\top} \bfH \bfv = 
    \begin{bmatrix}
        0 \\ \bfR \bfv
    \end{bmatrix}, 
\end{align}
where $\bfH = \begin{bmatrix} \boldsymbol{0}_{3 \times 1} & \bfI_3 \end{bmatrix}^{\top} \in \mathbb{R}^{4 \times 3}$ and $\bfR \in SO(3)$ is a rotation matrix achieving the same transformation as $\bfq$. Quaternions form a Lie group $\bbH_{1}$ with associated Lie algebra $\mathfrak{h}_{1}$. A quaternion $\bfq \in \bbH_{1}$ is related to a Lie algebra element $\bfphi^{\wedge} \in \mathfrak{h}_1$ through the exponential $\exp : \mathfrak{h}_{1} \rightarrow \bbH_1$ and logarithm $\log : \bbH_1 \rightarrow \mathfrak{h}_{1}$ maps: 
\begin{align}
    \bfq &= \exp{\left( \bfphi^{\wedge} \right)} = \begin{bmatrix}
        \cos{\left( \| \bfphi \|/2 \right)} \\ 
        \sin{\left( \| \bfphi \|/2  \right) \frac{\bfphi}{\|\bfphi\|}}
    \end{bmatrix},\\
    \bfphi^{\wedge} &= \frac{\bfphi}{2} = \log{\left( \bfq \right)} = \operatorname{atan2}{\left( \| \bfq_{v} \|, q_{w} \right)} \frac{\bfq_{v}}{\| \bfq_{v} \|},
\end{align}
where the division by $2$ accounts for the double cover of the space of rotation $SO(3)$ since $\bfq$ and $-\bfq$ correspond to the same rotation. A quaternion $\bfq$ is perturbed with an axis-angle vector $\bfphi \in \mathbb{R}^{3}$ as: 
\begin{align}
    \bfq^{+} = \bfq \otimes \exp{\left( \bfphi / 2 \right)}.
\end{align}
For small perturbations $\| \bfphi \| < \epsilon$, 
\begin{align}
    \bfq^{+} \approx \bfq + \frac{1}{2} \bfQ(\bfq) \bfphi, \quad \bfQ(\bfq) = \left[ \bfq \right]_{L} \bfH \in \mathbb{R}^{4 \times 3},
\end{align}
where $\bfQ(\bfq)$ is the orientation Jacobian. Please refer to \cite{sola2017quaternion, sola2018micro, jackson2021planning} for further details.

\section{PROBLEM STATEMENT}
\label{sec:problem_statement}

Consider a robot with state $\bfx_k \triangleq (\bfp_{k}, \bfq_{k}, \bfv_{k}, \bfomega_{k})$ at time $t_k$, consisting of position $\bfp_{k} \in \mathbb{R}^{3}$, orientation $\bfq_k \in \bbH_{1}$, body linear velocity $\bfv_k \in \mathbb{R}^{3}$, and body angular velocity $\bfomega_k \in \mathbb{R}^{3}$. We describe the evolution of the robot state with a discrete-time dynamics model
\begin{align} \label{eq:dynamics_model}
    \bfx_{k+1} = \bff_{\bftheta}(\bfx_k, \bfu_k),
\end{align}
where $\bfu_k \in \bbR^m$ is the control input and $\bff_{\bftheta}$ is a neural network parametrized by $\bftheta \in \mathbb{R}^{d}$. For instance, in the case of a quadrotor robot, the control input is $m=4$ dimensional, consisting of scalar thrust and 3D torque. Given a dataset $\calD = \{ \bfx_{0:T}^{(n)}, \bfu_{0:T}^{(n)} \}_{n=1}^{N}$ of state-control trajectories, our objective is to determine model parameters $\bftheta$ that minimize:
\begin{align} \label{eq:model_learning_problem}
    \begin{alignedat}{1}
    &\min_{\bftheta}\; r(\bftheta) + \sum_{n=1}^{N} \sum_{k=0}^{T} \ell(\bar{\bfx}_{k}^{(n)}, \bfx_{k}^{(n)}) \\ 
    &\text{s.t.} \quad \bar{\bfx}_{k+1}^{(n)} = \bff_{\bftheta}(\bar{\bfx}_k^{(n)}, \bfu_k^{(n)}), \quad \bar{\bfx}_{0}^{(n)} = \bfx_{0}^{(n)},
    \end{alignedat} 
\end{align}
where $\ell(\bar{\bfx}, \bfx)$ measures the error between a predicted state $\bar{\bfx}$ and an actual state $\bfx$. The term $r(\bftheta)$ regularizes the network parameters $\bftheta$. 

Once the dynamics model $\bff_{\bftheta}$ is estimated, it can be used to design a control policy $\bfu_{k} = \bfpi(\bfx_{k}, \bfx_{k}^{\star}, \bftheta)$ for the robot to track a target reference trajectory $\bfx_{k}^{\star}$ for $k = t, \ldots, t+T$.

Training the dynamics model as in \eqref{eq:model_learning_problem} from an uninformed initialization of the parameters $\bftheta$ may take a long time.  We consider the case where the model is initialized with pretrained parameters $\bftheta$ offline but then needs to be adapted online quickly in response to unmodeled disturbances affecting the evolution of the states $\bfx_k$. Updating the model parameters online, in turn, affects the control policy $\bfpi$ and can be used to achieve disturbance rejection and robust tracking.

\section{TECHNICAL APPROACH}
\label{sec:technical_approach}
\begin{figure}[!t]
    \centering
    \includegraphics[width=\linewidth]{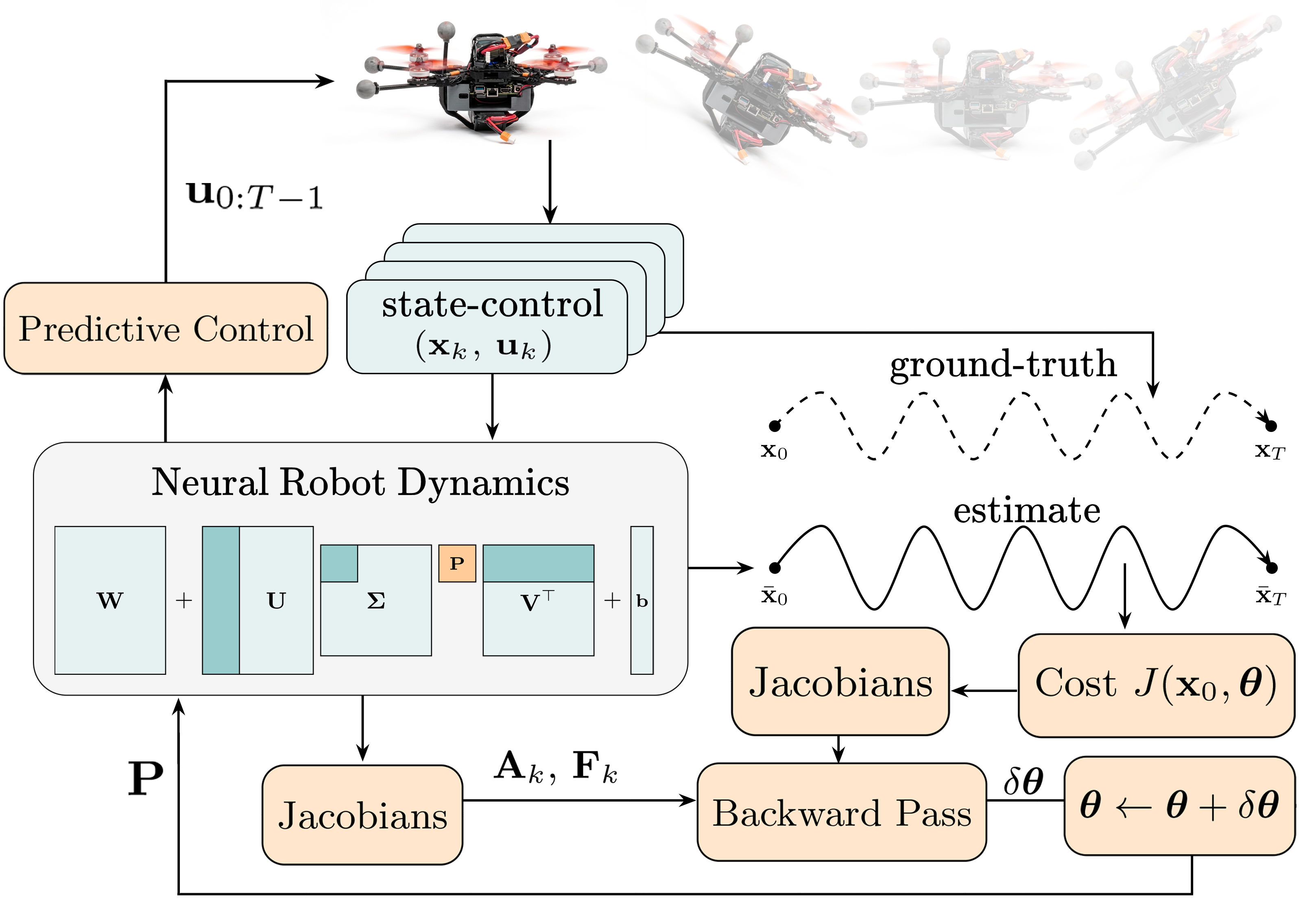}
    \caption{Overview of our approach for on-the-fly neural dynamics learning and predictive control.}
    \label{fig:overview}
\end{figure}

We present our approach in three stages: first, offline training of a prior dynamics model in Sec.~\ref{sec:dynamics_learning}, followed by online adaptation of learned model in Sec.~\ref{sec:online_model_learning}, and finally, predictive control in Sec.~\ref{sec:predictive_control}. See Fig.~\ref{fig:overview} for an overview of the system.  

\subsection{Offline Incremental Dynamics Learning} \label{sec:dynamics_learning}
It is common to learn a dynamics model $\bff_{\bftheta}$ that directly predicts the next state $\bfx_{k+1}$ as in Eq.~\eqref{eq:dynamics_model}. However, this expects a neural network to predict values far from zero and to implicitly satisfy the unit norm quaternion constraint. Instead, we formulate an incremental dynamics model $\delta \bff_{\bftheta}(\bfx_k, \bfu_k)$ that predicts the state change $\delta \bfx_{k} = \begin{bmatrix} \delta \bfp_{k}^{\top} & \delta \bfphi_{k}^{\top} & \delta \bfv_{k}^{\top} & \delta \bfomega_{k}^{\top} \end{bmatrix}^{\top} \in \mathbb{R}^{12}$, where $\delta \bfphi \in \mathbb{R}^{3}$ is a local orientation perturbation in $\mathfrak{h}_1$.
We define it as 
\begin{align}
    \label{eq:discrete_time_incremental_dynamics}
    \bfx_{k+1} = \bfx_{k} \oplus \delta \bfx_k =  \bfx_{k} \oplus \delta \bff_{\bftheta}(\bfx_k, \bfu_k), 
\end{align}
where $\oplus$ emphasizes that some additions are over the $\bbH_{1}$ manifold, defined as follows:
\begin{equation}
    \begin{aligned}
         \bfp_{k+1} &= \bfp_{k} + \delta \bfp_{k}, \quad \bfq_{k+1} = \bfq_k \otimes \exp{\left( \delta \bfphi_{k}^{\wedge} \right)}, \\ 
         \bfv_{k+1} &=  \bfv_{k} + \delta \bfv_{k}, \quad \bfomega_{k+1} = \bfomega_{k} + \delta \bfomega_{k}.
    \end{aligned}
\end{equation} 
With this design choice, we avoid forcing the network to implicitly learn the unit norm constraint of quaternions and guarantee that the resulting quaternion, by construction, remains on $\bbH_{1}$. The position information can be omitted in the incremental dynamics function since the dynamics of physical robot systems are translational-invariant. Thus, we end up with 
\begin{align}
    \delta \bff_{\bftheta}: (\bfq_{k}, \bfv_{k}, \bfomega_{k}, \bfu_{k}) \mapsto (\delta \bfp_{k}, \delta \bfphi_{k}, \delta \bfv_{k}, \delta \bfomega_{k}). 
\end{align}

We learn the parameters $\bftheta$ offline from a dataset $\calD$ of state-control trajectories, as mentioned in Sec.~\ref{sec:problem_statement}. For each trajectory, we roll out the model in  Eq.~\eqref{eq:discrete_time_incremental_dynamics} from initial state $\bfx_{0}^{(n)}$ under control inputs $\bfu_{0:T-1}^{(n)}$ to obtain predicted states $\bar{\bfx}_{1:T}^{(n)}$. Training minimizes the multi-step prediction error between predicted and ground-truth state via the loss function
\begin{align} \label{eq:state_cost}
    \ell(\bar{\bfx}_k^{(n)}, \bfx_k^{(n)}) = \| \bfe_{k}^{(n)} \|_{\bfQ}^{2}, \quad \bfe_{k}^{(n)} =  \bar{\bfx}_{k}^{(n)} \ominus \bfx_{k}^{(n)}, 
\end{align}
with $\ominus$ defined as: 
\begin{equation}
    \begin{aligned}
        \bfe_{\bfp} &= \bar{\bfp}_{k}^{(n)} - \bfp_k^{(n)}, \quad \bfe_{\bfq} = 2 \log{\left( (\bfq_k^{(n)})^{-1} \otimes \bar{\bfq}_{k}^{(n)} \right)} \\ 
        \bfe_{\bfv} &= \bar{\bfv}_{k}^{(n)} - \bfv_k^{(n)}, \quad \bfe_{\bfomega} = \bar{\bfomega}_{k}^{(n)} - \bfomega_k^{(n)}.
    \end{aligned}
\end{equation}
To give equal influence to the loss components, the weight matrix $\bfQ$ is set to the inverse covariance of the training dataset statistics. For robustness, we perturb the initial state $\bfx_{0}^{(n)}$ at each iteration with small additive Gaussian noise $\bfepsilon \sim \calN\left( \boldsymbol{0}, \sigma^2 \bfI \right)$. For quaternions, we perturb with $\bfq_{0} \otimes \exp{(\bfepsilon_{\bfq}^{\wedge})}$ so that perturbed quaternions remain on $\bbH_{1}$. Inspired by \cite{xu2025neural}, we apply input-output normalization to avoid bias and dominance of large-magnitude states in training. We normalize the inputs with physics-based limits to make them invariant to the training distribution as: 
 \begin{align}
     \check{\bfv}_{k}^{(n)} &= \frac{\bar{\bfv}_{k}^{(n)}}{v_{\text{max}}}, \quad \check{\bfomega}_{k}^{(n)} = \frac{\bar{\bfomega}_{k}^{(n)}}{\omega_{\text{max}}}, \quad  
     \check{\bfu}_{k}^{(n)} = \frac{\bar{\bfu}_{k}^{(n)} - u_{\text{max}}/2}{u_{\text{max}}/2}, \notag 
 \end{align}
 while the outputs are standardized using the mean and standard deviation statistics of the dataset. We represent the incremental dynamics $\delta \bff_{\bftheta}$ with a multi-layer perceptron (MLP) neural network with three hidden-layers of width $64$. Each layer $l$ has weight matrix $\bfW^{(l)}$, bias $\bfb^{(l)}$, and $\tanh(\cdot)$ activation function. The training is carried out using JAX \cite{bradbury2018jax} with Adam optimizer \cite{loshchilovdecoupled} to iteratively optimize the parameters $\bftheta = \{\bfW^{(l)},\bfb^{(l)}\}_l$.

\subsection{Online Adaptation of Learned Model} \label{sec:online_model_learning}
Fine-tuning all model parameters online is computationally expensive for real-time operation. Inspired by \cite{balazy2025lora}, we update a subset of parameters corresponding to the dominant singular vector directions of each weight matrix $\bfW^{(l)}$ in the MLP. For each layer $l$, the adapted weight matrix at iteration $i$ is defined as:  
\begin{equation*}
    \bfW^{(l)}_{i} = \bfW^{(l)} + \delta \bfW^{(l)}_{i} = \bfW^{(l)} + \bfU^{(l)} \bfSigma^{(l)} \bfP_{i}^{(l)} \bfV^{(l) \top}, 
\end{equation*}
where $\bfU^{(l)} \in \mathbb{R}^{m \times p}$, $\bfV^{(l)} \in \mathbb{R}^{n \times p}$, and $\bfSigma^{(l)} \in \mathbb{R}^{p \times p}$ are the left singular vectors, right singular vectors, and dominant singular values with $p \ll \min(m, n)$, respectively, obtained from a rank-$p$ truncated singular value decomposition (SVD) of the pretrained $\bfW^{(l)}$ in Sec.~\ref{sec:dynamics_learning}. The sole tunable matrix is $\bfP_{i}^{(l)} \in \mathbb{R}^{p \times p}$ and is initially set to zero, while the pretrained matrices $\bfW^{(l)}$, $\bfU^{(l)}$, $\bfV^{(l)}$, and $\bfSigma^{(l)}$ remain frozen throughout online adaptation. In the remainder of the paper, \NEW{we reserve $\bftheta_{0}$ for the frozen pretrained parameters and let $\bfeta_{i} \in \bbR^{Lp^{2}}$ be the collected tunable parameters $\bfP_{i}^{(l)}$ for $l = 1, \ldots, L$ in a single vector. Since the pretrained matrices are frozen, the adapted parameters are a function $\bftheta(\bfeta)$ of $\bfeta$ alone. We denote the resulting adapted dynamics model by $\bff_{\bftheta(\bfeta)}$, which recovers the pretrained model $\bff_{\bftheta_{0}}$ at $\bfeta = \boldsymbol{0}$.}

During online adaptation, we consider a single trajectory in Eq.~\eqref{eq:model_learning_problem}, omitting superscript $n$. The nonlinear trajectory optimization problem is solved iteratively by quadraticizing the cost and linearizing the dynamics about a nominal state-control trajectory. Each iteration consists of two passes. The forward pass rolls out the nonlinear dynamics \NEW{$\bff_{\bftheta(\bfeta)}$ from the current low-rank parameters $\bfeta$}, producing an updated nominal state-control trajectory $(\bar{\bfx}_{0:T}, \bar{\bfu}_{0:T-1})$ that is, by construction,  dynamically feasible. The backward pass, then, solves a Gauss-Newton approximation of Differential Dynamic Programming \cite{tassa2012synthesis} via a Riccati recursion to update the \NEW{low-rank parameters $\bfeta$}. This process repeats until convergence. Thus, the optimal \NEW{low-rank} parameters are obtained as: 
\begin{align} \label{eq:optimal_parameters}
    \NEW{\bfeta^{*} = \argmin_{\bfeta} J(\bfx_{0}, \bfeta),} 
\end{align}
\NEW{where $J(\bfx_{0}, \bfeta) := r(\bfeta) + \sum_{k=0}^{T} \ell(\bar{\bfx}_{k}, \bfx_{k})$}. At each step $k$, we approximate the cost-to-go \NEW{$J_{k} := J(\bfx_k, \bfeta)$} by a second-order Taylor expansion around a nominal trajectory with $\bfx_k := \bar{\bfx}_k + \delta \bfx_k$ and \NEW{$\bfeta := \bar{\bfeta} + \delta \bfeta$} as follows:
\NEW{
\begin{equation} \label{eq:cost_to_go_approximation}
        J_{k} \!\approx\! \bar{J}_k + 
        \begin{bmatrix}
            \bfJ^{\bfx}_{k} \\ \bfJ^{\bfeta}_{k}
        \end{bmatrix}^{\top}\!\! 
        \begin{bmatrix}
            \delta \bfx_k \\ \delta \bfeta
        \end{bmatrix}  
        + \frac{1}{2}\! 
        \begin{bmatrix}
            \delta \bfx_k \\ \delta \bfeta
        \end{bmatrix}^{\top}\!\!
        \begin{bmatrix}
            \bfJ^{\bfx \bfx}_{k} & \bfJ^{\bfx \bfeta}_{k} \\ 
            \bfJ^{\bfeta \bfx}_{k} & \bfJ^{\bfeta \bfeta}_{k}
        \end{bmatrix}
        \begin{bmatrix}
            \delta \bfx_k \\ \delta \bfeta
        \end{bmatrix} 
\end{equation}
}with
\NEW{
\begin{equation} \label{eq:ddp_backward_pass}
    \begin{aligned}
        \bar{J}_k &= J(\bar{\bfx}_k, \bar{\bfeta}), \\ 
        \bfJ^{\bfx}_{k} &= \bfell_{k}^{\bfx} + \bfA_k^{\top} \bfJ^{\bfx}_{k+1}, \\ 
        \bfJ^{\bfeta}_{k} &= \bfr_{k}^{\bfeta} + \bfF_k^{\top} \bfJ^{\bfx}_{k+1} + \bfJ^{\bfeta}_{k+1}, \\ 
        \bfJ^{\bfx \bfx}_{k} &= \bfell_{k}^{\bfx \bfx} + \bfA_k^{\top} \bfJ^{\bfx \bfx}_{k+1} \bfA_k, \\ 
        \bfJ^{\bfeta \bfeta}_{k} &= \bfr_{k}^{\bfeta \bfeta} + \bfF_k^{\top} \bfJ^{\bfx \bfx}_{k+1} \bfF_k + 2 \bfF_k^{\top} \bfJ^{\bfx \bfeta}_{k+1} + \bfJ^{\bfeta \bfeta}_{k+1}, \\ 
        \bfJ^{\bfx \bfeta}_{k} &= \bfA_k^{\top} \bfJ^{\bfx \bfx}_{k+1} \bfF_k + \bfA_k^{\top} \bfJ^{\bfx \bfeta}_{k+1}.
    \end{aligned}
\end{equation}
}where \NEW{$\bfr_{k}^{\bfeta}$ and $\bfr_{k}^{\bfeta \bfeta}$} are zero except at $k=T$. The matrices $\bfA_k$ and $\bfF_k$ denote the dynamics Jacobians with respect to state and \NEW{low-rank} parameters, respectively, and can be obtained using automatic differentiation, e.g., via LibTorch \cite{paszke2019pytorch}, as
\NEW{
\begin{equation} \label{eq:Jacobians}
    \begin{aligned}
        \bfA_k &= \bfE(\bar{\bfq}_{k+1})^{\top} \left.\frac{\partial \bff_{\bftheta(\bfeta)}(\bfx, \bfu)}{\partial \bfx} \right\vert_{(\bar{\bfx}_k, \bar{\bfu}_k, \bar{\bfeta})} \bfE(\bar{\bfq}_{k}), \\ 
        \bfF_{k} &= \bfE(\bar{\bfq}_{k+1})^{\top} \left.\frac{\partial \bff_{\bftheta(\bfeta)}(\bfx, \bfu)}{\partial \bfeta} \right\vert_{(\bar{\bfx}_k, \bar{\bfu}_k, \bar{\bfeta})} 
    \end{aligned}    
\end{equation}
}where $\bfE(\bfq) = \operatorname{diag}(\bfI_3, \bfQ(\bfq), \bfI_3, \bfI_3)$ is a tangent space projection mapping to the three-dimensional axis-angle representation. For our neural network, these matrices admit closed-form expressions, 
\begin{equation}
\begin{aligned}
    \NEW{\frac{\partial \bff_{\bftheta(\bfeta)}(\bfx, \bfu)}{\partial \bfx}} &= \frac{\partial \bfx_{k+1}}{\partial \bfx_{k}} + \frac{\partial \bfx_{k+1}}{\partial \delta \bfx_{k}} \frac{\partial \delta \bfx_{k}}{\partial \bfz_{k}} \frac{\partial \bfz_{k}}{\partial \bfx_{k}} \\  
    \NEW{\frac{\partial \bff_{\bftheta(\bfeta)}(\bfx, \bfu)}{\partial \bfu}} &= \frac{\partial \bfx_{k+1}}{\partial \bfu_{k}} + \frac{\partial \bfx_{k+1}}{\partial \delta \bfx_{k}} \frac{\partial \delta \bfx_{k}}{\partial \bfz_{k}} \frac{\partial \bfz_{k}}{\partial \bfu_{k}} \\ 
    \NEW{\frac{\partial \bff_{\bftheta(\bfeta)}(\bfx, \bfu)}{\partial \bfeta}} &= \frac{\partial \bfx_{k+1}}{\partial \delta \bfx_{k}} \NEW{\frac{\partial \delta \bfx_{k}}{\partial \delta \bfeta}}
\end{aligned}
\end{equation}
with
\begin{equation}
\begin{aligned}
    \frac{\partial \bfx_{k+1}}{\partial \bfx_{k}} &= \operatorname{diag}(\bfI_{3}, \, \left[ \exp{\left( \delta \bfphi_{k}^{\wedge} \right)} \right]_{R}, \, \bfI_3, \, \bfI_3 ) \\ 
    \frac{\partial \bfx_{k+1}}{\partial \delta \bfx_{k}} &= \operatorname{diag}( \bfI_3, \, \frac{1}{2}\, \left[ \bfq_{k} \right]_{L} \bfH, \, \bfI_3, \, \bfI_3 ) \\ 
    \frac{\partial \delta \bfx_{k}}{\partial \bfz_{k}} &= \bfW^{(L)}\, \bfD^{(L)} \cdots \bfD^{(1)} \bfW^{(1)}, \\ 
    \bfD^{(l)} &= \operatorname{diag}(1 - \tanh{(\bfh^{(l)})}^{2}),
\end{aligned}
\end{equation}
where $\bfz_k = (\bfq_{k}, \bfv_{k}, \bfomega_{k}, \bfu_{k})$ denote the input to the neural network and $\bfh^{(l)} = \bfW^{(l)} \bfh^{(l-1)} + \bfb^{(l)}$ is the pre-activation at layer $l$ with $\bfh_{0} = \bfz_{k}$. The partial derivatives $\frac{\partial \bfz_{k}}{\partial \bfx_{k}}$ and $\frac{\partial \bfz_{k}}{\partial \bfu_{k}}$ represent selection matrices such that $\bfz_{k} = \frac{\partial \bfz_{k}}{\partial \bfx_{k}} \bfx_{k} + \frac{\partial \bfz_{k}}{\partial \bfu_{k}} \bfu_k$. We omit the closed-form expression for the parameter Jacobian for conciseness. 

Substituting the quadratic approximation in Eq.~\eqref{eq:cost_to_go_approximation} into Eq.~\eqref{eq:optimal_parameters} and neglecting non-dependent terms leads to 
\NEW{
\begin{align} 
    \min_{\delta \bfeta} \; \bfJ^{\bfeta \top}_{0} \delta \bfeta + \delta \bfx_{0}^{\top} \bfJ^{\bfx \bfeta}_{0} \delta \bfeta 
    &+ \frac{1}{2} \delta \bfeta^{\top} \bfJ^{\bfeta \bfeta}_{0} \delta \bfeta. 
\end{align}
}Solving the minimization above yields the optimal \NEW{low-rank parameters increment $\delta \bfeta$:}
\NEW{
\begin{align}
    \delta \bfeta^* = -\left( \bfJ^{\bfeta \bfeta}_{0} \right)^{-1} \bfJ^{\bfeta}_{0}. 
\end{align}
}To ensure that \NEW{$\bfJ^{\bfeta \bfeta}_{0}$} is positive semidefinite, we regularize it iteratively with \NEW{$\bfJ^{\bfeta \bfeta}_{0} \gets \bfJ^{\bfeta \bfeta}_{0} + \mu\, \bfI$} with $\mu > 0$. Also, to ensure cost-to-go reduction, we use line search on the \NEW{low-rank parameters $\bar{\bfeta}^{\prime} = \bar{\bfeta} + \alpha\, \delta \bfeta$} with $\alpha \in [0, 1]$ to determine the update amount and evaluate the cost-to-go \NEW{$J(\bar{\bfx}_{0}^{\prime}, \bar{\bfeta}^{\prime})$} on $\bar{\bfx}_{0:T}^{\prime}$ obtained from \NEW{$\bff_{\bftheta(\bar{\bfeta}^{\prime})}$}. We summarize the steps in Algorithm~\ref{alg:ddp}.

\subsection{Predictive Control} \label{sec:predictive_control}
Given the dynamics model learned in Sec~\ref{sec:online_model_learning}, we formulate a tracking control policy $\bfu_{k} = \bfpi(\bfx_{k}, \bfx_{k}^{\star}, \NEW{\bftheta(\bfeta)})$ as a finite-horizon optimal control problem:
\begin{equation}
\begin{aligned} \label{eq:optimal_control_problem}
    \begin{alignedat}{1}
    &\min_{\bfu_{0:T-1}} \; V(\bfx_{0}, \bfu_{0:T-1}) :=  \sum_{k=0}^{T-1} \ell_{k}(\bar{\bfx}_{k}, \bfx_{k}^{\star}) + c_{k}(\bar{\bfu}_{k}, \bfu_{k})\\ 
    &\NEW{\text{s.t.} \quad \bar{\bfx}_{k+1} = \bff_{\bftheta(\bfeta)}(\bar{\bfx}_k, \bfu_k), \quad \bar{\bfx}_{0} = \bfx_{0},}
    \end{alignedat}  
\end{aligned}
\end{equation}
where $c_{k}(\bar{\bfu}_{k}, \bfu_{k})$ is the control cost and $\ell_{k}(\bar{\bfx}_{k}, \bfx_{k}^{\star})$ is the state cost in Sec.~\ref{sec:online_model_learning} but with reference $\bfx_{k}^{\star}$ instead of actual $\bfx_{k}$. Analogous to \eqref{eq:ddp_backward_pass}, we end up with the following backward pass: 
\begin{equation}
    \begin{aligned}
        \bfV_{k}^{\bfx} &= \bfell_{k}^{\bfx} + \bfA^{\top}_k \bfV_{k+1}^{\bfx}, \qquad \quad \; 
        \bfV_{k}^{\bfu} = \bfc_{k}^{\bfu} + \bfB^{\top}_k \bfV_{k+1}^{\bfx} \\
        \bfV_{k}^{\bfx \bfx} &= \bfell_{k}^{\bfx \bfx} + \bfA_k^{\top} \bfV_{k+1}^{\bfx \bfx} \bfA_k, \quad    
        \bfV_{k}^{\bfx \bfu} =  \bfA_k^{\top} \bfV_{k+1}^{\bfx \bfx} \bfB_k \\ 
        \bfV_{k}^{\bfu \bfu} &= \bfc_{k}^{\bfu \bfu} + \bfB_k^{\top} \bfV_{k+1}^{\bfx \bfx} \bfB_k 
    \end{aligned}
\end{equation}
with $\bfA_{k}$ from Eq.~\eqref{eq:Jacobians} and the Jacobian with respect to the \NEW{input: 
\begin{align}
    \bfB_{k} &= \bfE(\bar{\bfq}_{k+1})^{\top} \left.\frac{\partial \bff_{\bftheta(\bfeta)}(\bfx, \bfu)}{\partial \bfu} \right\vert_{(\bar{\bfx}_k, \bar{\bfu}_k, \bar{\bfeta})}. 
\end{align}
}The optimal control \NEW{$\bfu_{k} = \bfpi(\bfx_{k}, \bfx_{k}^{\star}, \bftheta(\bfeta))$} is 
\begin{equation} \label{eq:optimal_control}
    \begin{aligned}
        \bfu_k &= \bar{\bfu}_k + \delta \bfu_k, &\hfill \delta \bfu_k &= \bfk_k + \bfK_k \delta \bfx_k \\ 
        \bfk_k &= - (\bfV_k^{\bfu \bfu})^{-1} \bfV_k^{\bfu}, &\hfill 
        \bfK_k &= - (\bfV_k^{\bfu \bfu})^{-1} \bfV_{k}^{\bfu \bfx}
    \end{aligned}
\end{equation}
where $\bar{\bfu}_k$ is the nominal control trajectory from the previous iteration and $\delta \bfx_k$ is the state deviation of the new nominal trajectory and the previous one. At each control cycle, we warm start from the time-shifted previous solution. 

\begin{algorithm}[t]
\caption{Online Adaptation of \NEW{Low-Rank Parameters $\bfeta$}}
\label{alg:ddp}
\begin{algorithmic}[1]
\Require $\bfx_{0:T}, \bfu_{0:T-1}, \NEW{\bfeta_{i}}$
\Require Linearization $\delta \bfx_{k+1} = \bfA_k \delta \bfx_k + \bfF_k \NEW{\delta \bfeta_{i}}$
\State Rollout $\bar{\bfx}_{0:T}$ with $(\bfx_{0}, \bfu_{0:T-1})$ using \NEW{$\bff_{\bftheta(\bfeta_{i})}$}
\State Compute \NEW{$J(\bfx_{0}, \bfeta_{i})$}
\State Let $\bfJ^{\bfx}_{T} = \bfell_{T}^{\bfx}, \; \NEW{\bfJ^{\bfeta}_{T} = \bfr_{T}^{\bfeta}}$, \; $\bfJ^{\bfx \bfx}_{T} = \bfell_{T}^{\bfx \bfx}$, \; $\NEW{\bfJ^{\bfeta \bfeta}_{T} = \bfr_{T}^{\bfeta \bfeta}}$
\For{$k = T-1$ to $0$}
  \State Compute $\bfA_k$ and $\bfF_k$ as in Eq.~\eqref{eq:Jacobians}
  \State Compute backward pass Jacobians as in Eq.~\eqref{eq:ddp_backward_pass}
\EndFor
\Ensure \NEW{$\bfJ^{\bfeta \bfeta}_{0} \succ 0$ else $\bfJ^{\bfeta \bfeta}_{0} \gets \bfJ^{\bfeta \bfeta}_{0} + \mu \bfI$ \quad $(\mu > 0)$}
\State Compute \NEW{$\delta \bfeta_i = -\left( \bfJ^{\bfeta \bfeta}_{0} \right)^{-1} \bfJ^{\bfeta}_{0}$} 
\Repeat 
    \State Rollout $\bar{\bfx}_{1:T}^{\prime}$ with \NEW{line search $\bfeta_{i} + \alpha\,\delta \bfeta_{i}$ \quad $(\alpha > 0)$}
\Until{\NEW{$J(\bfx_{0}, \bfeta_{i} + \alpha \delta \bfeta_{i}) < J(\bfx_{0}, \bfeta_{i})$}}
\State Update \NEW{$\bfeta_{i+1} = \bfeta_{i} + \alpha \delta \bfeta_{i}$}
\end{algorithmic}
\end{algorithm}

\begin{figure*}[!t]
    \centering
    \includegraphics[width=\linewidth]{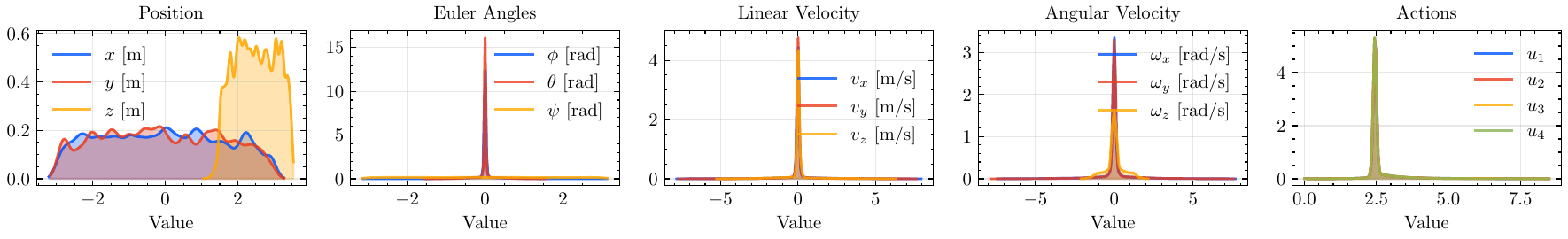}
    \includegraphics[width=\linewidth]{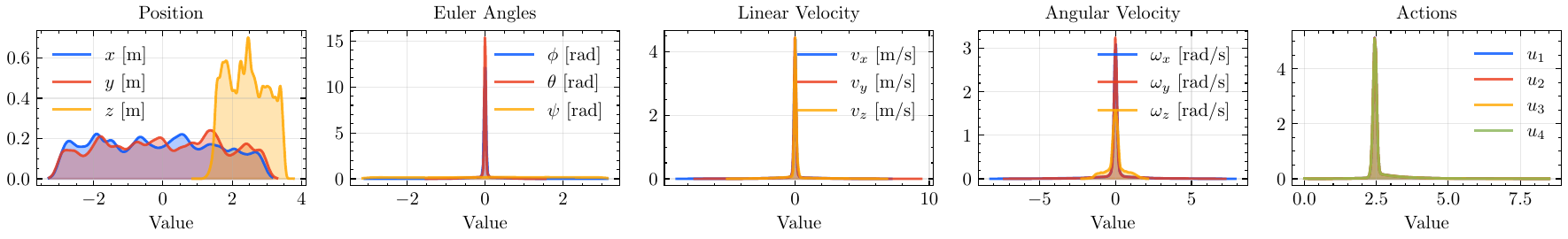}
    \caption{Range of values and their densities in the collected dataset of quadrotor positions, orientations, linear velocities, and angular velocities. The top row corresponds to the training set, and the bottom row to the validation set.}
    \label{fig:dataset_range}
\end{figure*}
\section{EVALUATION}
In this section, we first assess the prediction accuracy of the pretrained model from Sec.~\ref{sec:dynamics_learning}. Then, we evaluate the trajectory tracking performance of our learned model with and without online adaptation from Sec.~\ref{sec:online_model_learning} on a real quadrotor robot under changing conditions.

\subsection{Offline Pre-training}

We pretrained a quadrotor dynamics model using a dataset $\calD$ of state-control trajectories obtained from a simulated quadrotor dynamics model with mass $m = 1.0\,\mathrm{kg}$, inertia $\bfM = \operatorname{diag}(0.0025, 0.0025, 0.004)\,  \mathrm{kg}\,\mathrm{m}^2$, arm lengths $l = 0.125\,\mathrm{m}$, and dynamics:
\begin{equation*}
\begin{aligned}
    \dot{\bfp} &= \bfR \bfv, \quad 
    \dot{\bfq} = \frac{1}{2} \left[ \bfq \right]_{L} \bfH \bfomega, \quad \dot{\bfomega} = \bfM^{-1} \left(\bftau -\bfomega  \times \bfM \bfomega \right), \\
    \dot{\bfv} &= \frac{f}{m} \bfe_3 - \bfomega \times \bfv + \bfR^{\top} \bfg, \quad 
    \bfR = \bfH^{\top} \left[ \bfq \right]_{L} \left[ \bfq \right]_{R}^{\top} \bfH,
\end{aligned}
\end{equation*}
where $\bfg = \begin{bmatrix} 0 & 0 & -9.81 \end{bmatrix}^{\top} \in \bbR^{3}$ is the gravity acceleration, $\bfe_{3} = \begin{bmatrix} 0 & 0 & 1 \end{bmatrix}^{\top} \in \bbR^{3}$, and the thrust $f$ and torque $\bftau$ are obtained from the motor thrusts $\bfu \in \mathbb{R}^{4}$ as:
\begin{equation}
    \begin{bmatrix}
        f \\ \bftau
    \end{bmatrix} = 
    \begin{bmatrix}
        1 & 1 & 1 & 1 \\ 
        -\frac{l}{\sqrt{2}} & \frac{l}{\sqrt{2}} & -\frac{l}{\sqrt{2}} & \frac{l}{\sqrt{2}} \\ 
        -\frac{l}{\sqrt{2}} & \frac{l}{\sqrt{2}} & \frac{l}{\sqrt{2}} & -\frac{l}{\sqrt{2}} \\ 
        -k_{\tau} & -k_{\tau} & k_{\tau} & k_{\tau}
    \end{bmatrix} \bfu, 
\end{equation}
with torque constant $k_{\tau} = 0.016\ \mathrm{N}\, \mathrm{m}\, \mathrm{s}^{2}$. Using a Runge-Kutta numerical integrator \cite{dormand1980family}, we generated $1000$ trajectories, with $500$ used for training and $500$ for validation, of $700$ samples each sampled at $100\,\mathrm{Hz}$, with the quadrotor controlled via a geometric controller \cite{lee2010geometric}. Fig.~\ref{fig:dataset_range} shows the range of values in the collected dataset.

We partitioned each trajectory into non-overlapping windows of length $T=10$, corresponding to a prediction horizon of $0.1$ seconds. From the initial state $\bfx_{0}$ in each window, we rolled out the model $\bff_{\bftheta}$ for $T-1$ steps recursively to obtain the predicted states $\bar{\bfx}_{1:T}$. Because $\bfq$ and $-\bfq$ represent the same physical attitude, we mapped to the upper hemisphere $q_{w} \geq 0$ whenever $q_{w} < 0$ to avoid discontinuity of double cover \cite{brescianini2013nonlinear}. The resulting quaternion $\bfq_{k+1}$ was then normalized after composition to prevent numerical drift. We then computed the loss $\ell(\bar{\bfx}_k^{(n)}, \bfx_k^{(n)})$ over the rollout. We trained for $10,000$ epochs with a batch size of $8192$ and a prediction horizon $T=10$. We used the Adam optimizer \cite{loshchilovdecoupled} 
with cosine learning rate schedule from $10^{-3}$ to $10^{-5}$. We trained the neural network model on a desktop equipped with Intel Core i7 3.6GHz CPU and Nvidia RTX 3090 Ti GPU.

We evaluated the model $\bff_{\bftheta}$ with $T=50$ on the validation dataset, matching the intended control horizon, to test whether the model generalizes beyond the training horizon. 
To assess the prediction accuracy, we report the root mean square error (RMSE) in position, orientation, linear velocity, and angular velocity as follows: 
\begin{align}
    \text{RMSE} = \sqrt{\left( \frac{1}{N} \sum_{k=0}^{N-1} \| \bfe_k \|_{2}^{2} \right)}, \quad \bfe_k = \bar{\bfx}_k \ominus \bfx_k. 
\end{align}
We observed that over a prediction horizon of $0.5$ seconds, the errors on average were $0.06\,\mathrm{m}$, $0.10\,\mathrm{rad}$, $0.26\,\mathrm{m/s}$, $0.40\,\mathrm{rad/s}$ for position, orientation, linear velocity, and angular velocity, respectively.



\begin{figure*}[t]
    \centering
    \subcaptionbox{}{\includegraphics[width=0.24\linewidth]{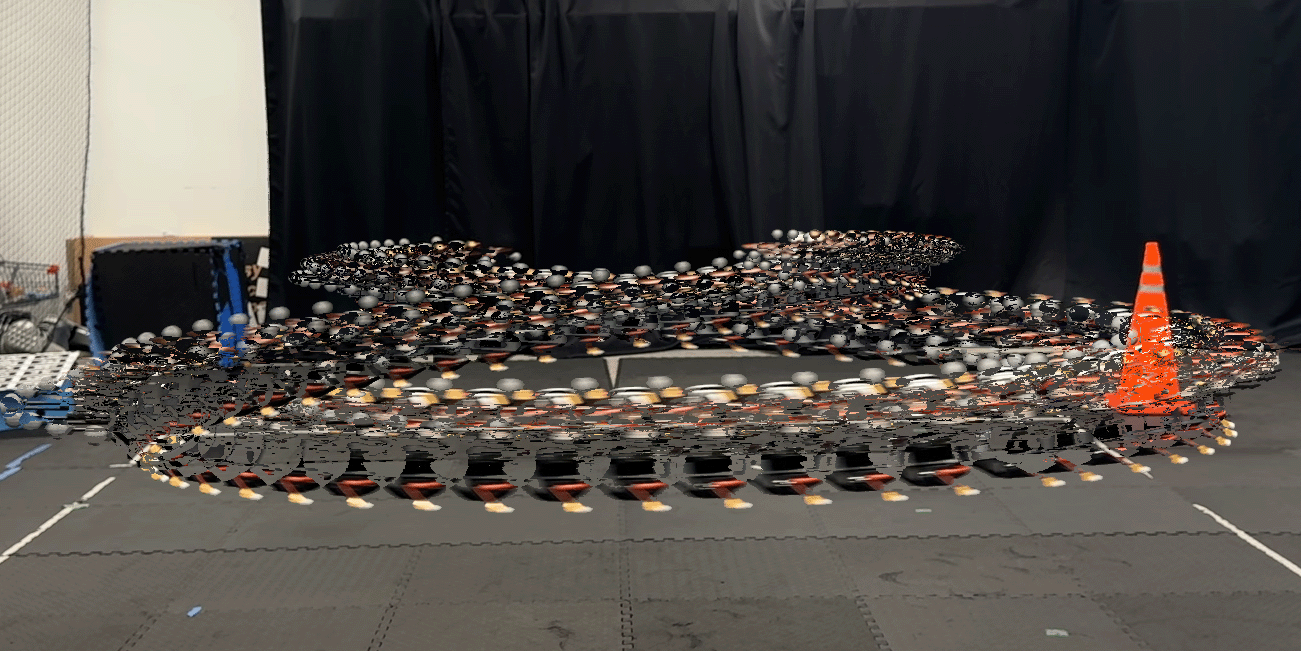}}%
    \subcaptionbox{}{\includegraphics[width=0.24\linewidth]{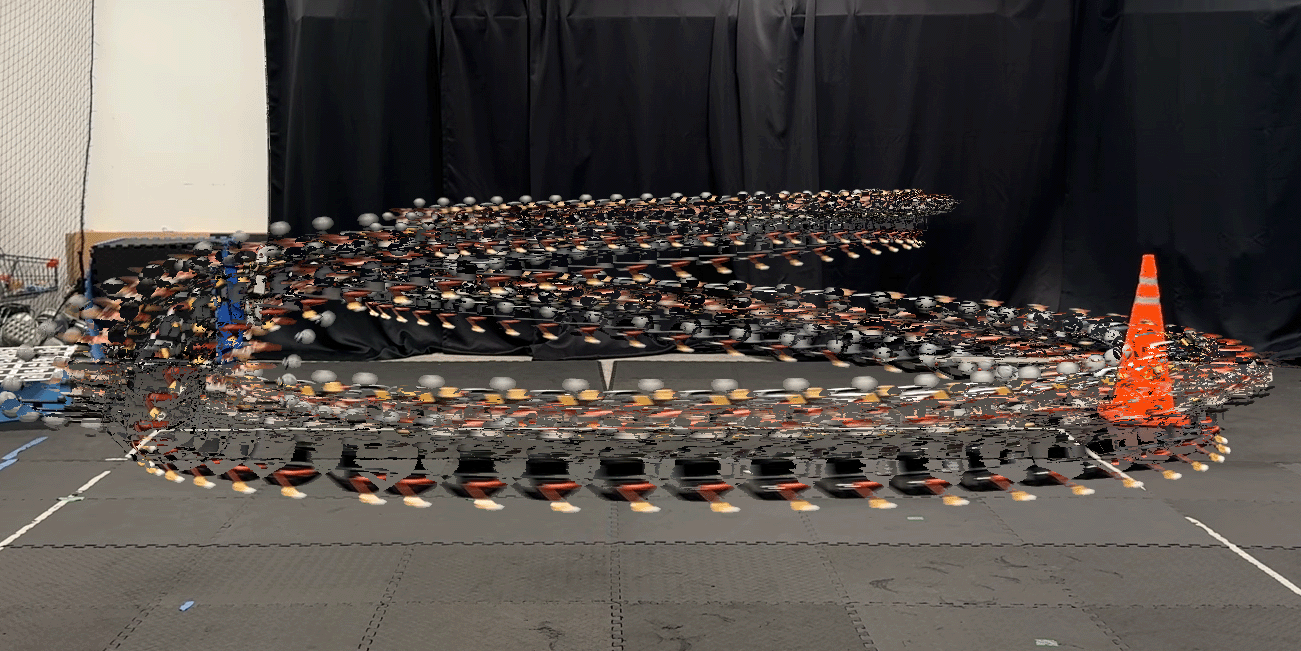}}%
    \subcaptionbox{}{\includegraphics[width=0.24\linewidth]{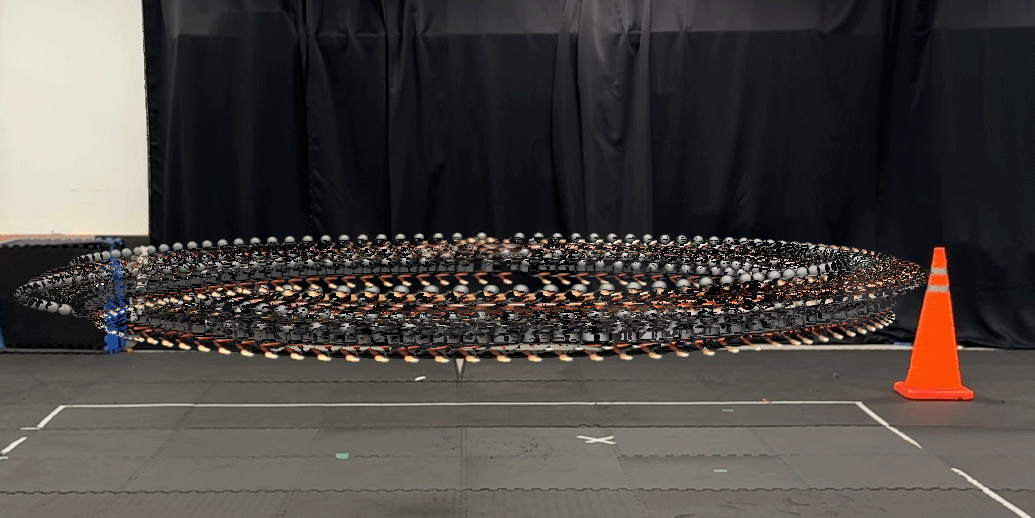}}%
    \subcaptionbox{}{\includegraphics[width=0.24\linewidth]{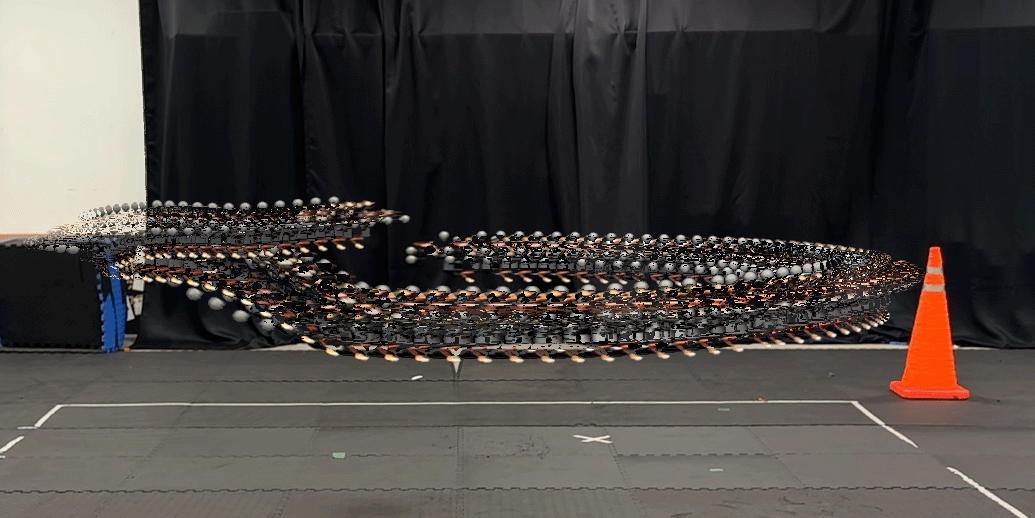}}
    \subcaptionbox{}{\includegraphics[width=0.5\linewidth]{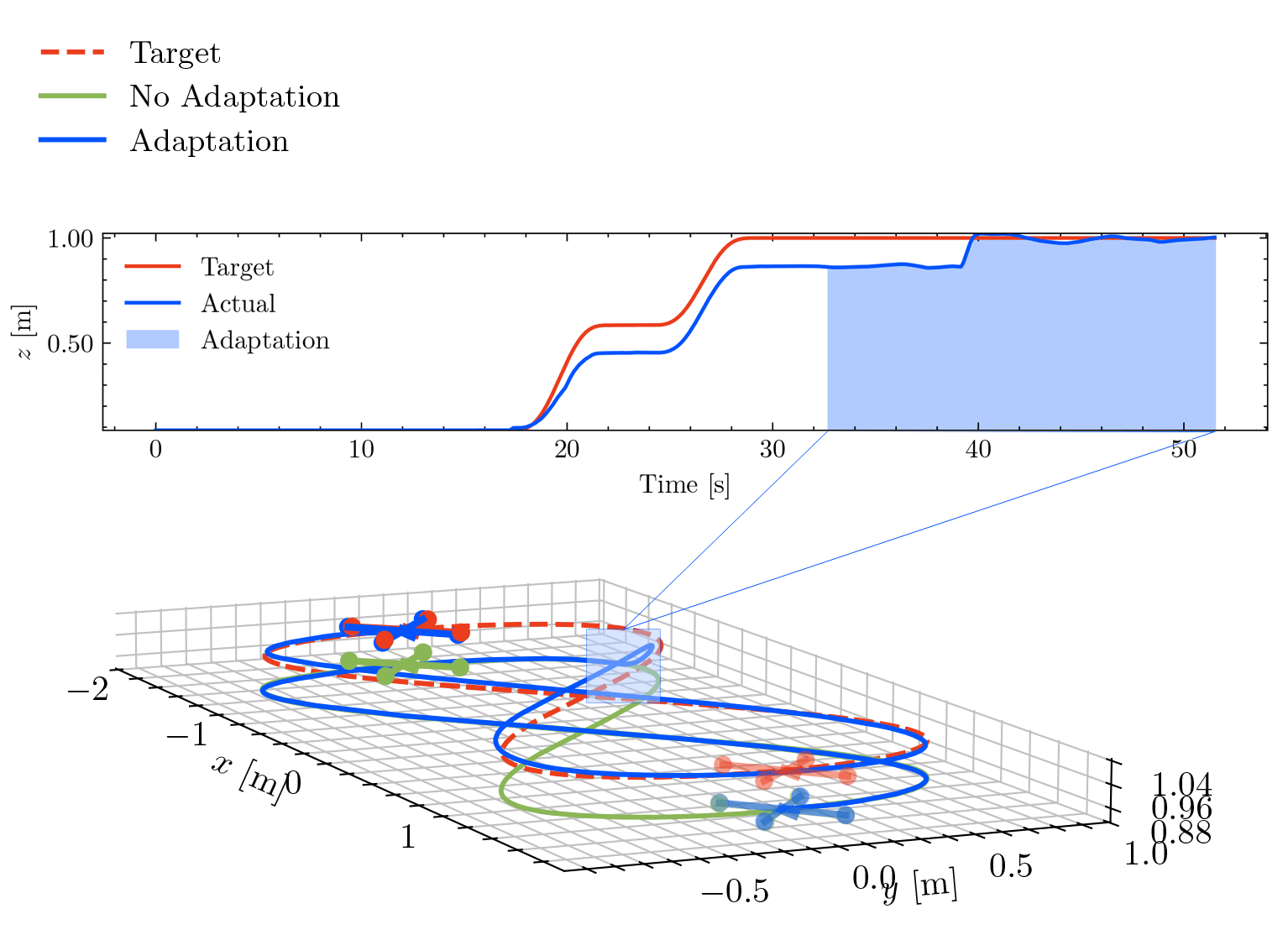}}%
    \hfill%
    \subcaptionbox{}{\includegraphics[width=0.5\linewidth]{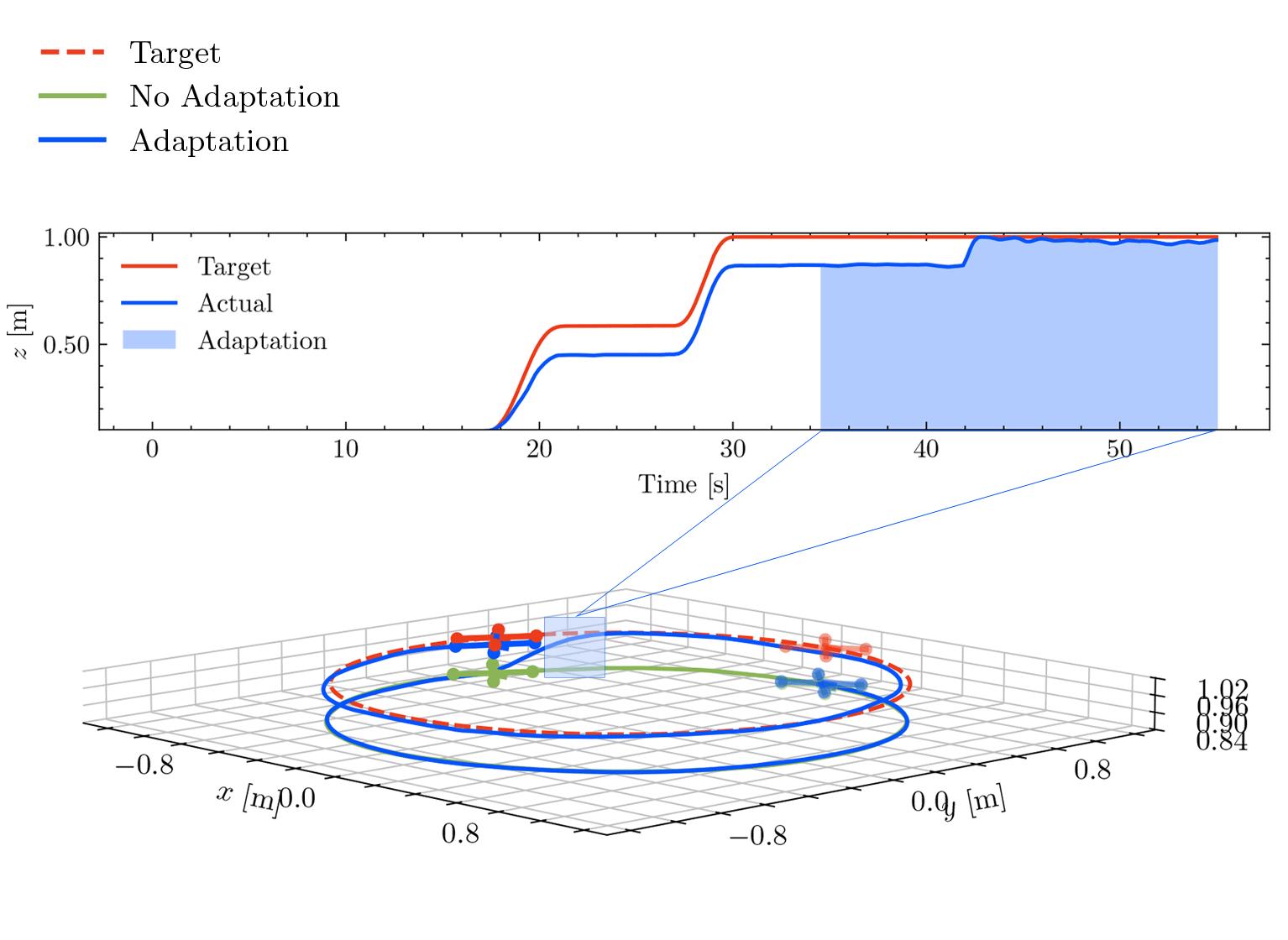}}
    \caption{Quadrotor tracking lemniscate and circular reference trajectories with an added $350\,\mathrm{g}$ payload ($35\%$ increase). (a–b) lemniscate without/with adaptation,  (c–d) circle without/with adaptation. In (e-f),  the transparent quadrotors denote the starting states while the bold ones denote the end states.}
    \label{fig:3d_plot}
\end{figure*}


\subsection{Real-World Online Adaptation}

We deployed the model pretrained in simulation on a real quadrotor and performed online model adaptation and predictive tracking control in Sec.~\ref{sec:technical_approach}. The quadrotor, shown in Fig.~\ref{fig:thumbnail}, is equipped with a microcontroller running a $32$-bit ARM Cortex-M7 at $216\,\mathrm{MHz}$, a BMI270 inertial measurement unit (IMU) delivering gyroscope measurements at $6.4\,\mathrm{kHz}$, and an Intel NUC on-board computer with a Core i7-8650U CPU. Motor commands are transmitted to the electronic speed controllers (ESCs) via the DShot protocol, enabling individual motor speed feedback. We augmented our predictive controller with incremental nonlinear dynamic inversion \cite{smeur2016adaptive} to provide robustness during real flights. The controller operates at $100\,\mathrm{Hz}$. Ground-truth pose data was obtained from an OptiTrack motion capture system at $240\,\mathrm{Hz}$. The method was implemented on-board using the open-source Agilicious flight stack \cite{foehn2022agilicious} with the C++ Eigen library for efficient matrix operations.

To evaluate adaptation under significant model mismatch, we attached an extra payload of $0.35\,\mathrm{kg}$ to the quadrotor, resulting in a $35\%$ increase in total mass. The quadrotor tracked lemniscate and circular reference trajectories shown in Fig.~\ref{fig:3d_plot}. For online adaptation, we set the rank to $p = 5$, reducing the number of adaptable parameters to $100$ (approximately $1\%$ of total number of neural network weights), while preserving expressiveness. Over each recorded state-control trajectory, we ran three iterations of Algorithm~\ref{alg:ddp}. The remaining parameters are listed in Table~\ref{tab:cost_weights}. Fig.~\ref{fig:3d_plot} shows a qualitative comparison of 3D trajectory tracking with and without adaptation. The adapted model rapidly compensates for the added payload and converges to the $1\,\mathrm{m}$ reference altitude within $\sim\,7$ seconds, while the non-adapted baseline remains below the reference altitude. Quantitatively, Table~\ref{tab:real_ate} reports the tracking RMSE in position and heading. With adaptation, position tracking improved by $21\%$ on the lemniscate trajectory and by $26\%$ on the circle trajectory. The heading errors remained comparable across both models, which is expected since the payload primarily affects the translational dynamics. 


\begin{table}[!htp]
\centering
\caption{Cost Weights}
\label{tab:cost_weights}
\scriptsize
\begin{tabular}{ll|lll}
\hline
\toprule
\multicolumn{2}{c|}{\textbf{Control}} & \multicolumn{2}{c}{\textbf{Adaptation}} & \\
\midrule
$\bfQ_{\bfp}$ 
& $\mathrm{diag}(200,\,200,\,200)$ 
& $\bfQ_{\bfp}$ 
& $\mathrm{diag}(10,\,10,\,10)$ 
& \\

$\bfQ_{\bfq}$ 
& $\mathrm{diag}(1.25,\,1.25,\,50)$ 
& $\bfQ_{\bfq}$ 
& $\mathrm{diag}(5,\,5,\,2)$ 
& \\

$\bfQ_{\bfv}$
& $\mathrm{diag}(1,\,1,\,1)$ 
& $\bfQ_{\bfv}$
& $\mathrm{diag}(1,\,1,\,1)$ 
& \\

$\bfQ_{\bfomega}$
& $\mathrm{diag}(1,\,1,\,1)$ 
& $\bfQ_{\bfomega}$
& $\mathrm{diag}(0.1,\,0.1,\,0.1)$ 
& \\

$\bfQ_{\bfu}$ 
& $2 \times \bfI_{4}$ 
& \NEW{$\bfQ_{\bfeta}$} 
& $0.1 \times \bfI_{p}$ 
& \\

$T$ 
& $50$ 
& $T$ 
& $50$ 
& \\
\bottomrule
\end{tabular}
\end{table}


\begin{table}[!htp]
\centering
\caption{Tracking performance reported in RMSE.}
\label{tab:real_ate}
\scriptsize
\setlength{\tabcolsep}{6pt}
\begin{tabular}{llccccc}
\toprule
\textbf{Trajectory} & \textbf{Method}
& \textbf{Position [$\mathrm{m}$]}
    & \textbf{Heading [$\mathrm{rad}$]}
    & \textbf{Overall} \\
\midrule
\multirow{2}{*}{Lemniscate}
    & No Adaptation & $0.136$ & $0.059$ & $0.196$  \\
    & With Adaptation & ${\bf 0.107}$ & ${\bf 0.058}$ & ${\bf 0.166}$ \\
\midrule
\multirow{2}{*}{Circle}
    & No Adaptation & $0.134$ & $0.063$ & $0.198$  \\
    & With Adaptation & ${\bf 0.099}$ & ${\bf 0.060}$ & ${\bf 0.159}$ \\
\bottomrule
\end{tabular}
\end{table}

\section{CONCLUSION}
We developed an approach for on-the-fly neural dynamics learning and predictive control for autonomous mobile robots. Our approach trains an incremental dynamics model offline and adapts it online using low-rank second-order parameter updates, reducing the tunable parameters to approximately $1\%$ of total parameters to enable real-time on-board execution. We demonstrated online model adaptation and control on a real quadrotor robot with extra $35\%$ payload, achieving better tracking performance compared to a non-adaptive baseline. Future work will focus on incorporating visual measurements instead of direct state observations and evaluating on a wider set of disturbance types.

\bibliographystyle{IEEEtran}
\bibliography{bib/references}

\end{document}